\documentclass[letterpaper, 10 pt, conference]{ieeeconf}  

\usepackage{tabularx} 
\usepackage{booktabs} 
\usepackage{graphicx} 
\usepackage{makecell} 

\usepackage{epsfig}
\usepackage{graphicx}
\usepackage{amsmath,epsfig}
\usepackage{graphicx}
\usepackage{textcomp}
\usepackage{xcolor}
\usepackage{rotating}
\usepackage{multirow}
\usepackage{booktabs}
\usepackage{colortbl}
\usepackage{overpic}
\usepackage{multirow}
\newcommand{\hsy}[1]{#1}
\definecolor{green}{RGB}{0,150,10}
\definecolor{blue}{RGB}{0,148,181}
\definecolor{orange}{RGB}{194,153,107}
\newcommand{\wb}[1]{#1}
\IEEEoverridecommandlockouts                              
\newcommand{\revision}[1]{#1}
\newcommand{\lingqiao}[0]

\newcommand{\delee}{\textit{Delta End-Effector}}
\newcommand{\deljoint}{\textit{Delta Joint Position}}
\newcommand{\absee}{\textit{Absolute End-Effector}}
\newcommand{\absjoint}{\textit{Absolute Joint Position}}

\newcommand{\GMP}{GMPs}

\title{\LARGE \bf

Effective Tuning Strategies for Generalist Robot Manipulation Policies

}

\author{Wenbo Zhang$^{1}$ Yang Li$^{2}$ Yanyuan Qiao$^{1}$ Siyuan Huang$^{3}$ Jiajun Liu$^{2}$  Feras Dayoub$^{1}$  Xiao Ma$^{4}$ Lingqiao Liu$^{1*}$
\thanks{$^1$ University of Adelaide, $^2$ Commonwealth Scientific and Industrial Research Organisation (CSIRO), $^3$ Shanghai Jiao Tong University, $^4$ TikTok}
\thanks{*Corresponding author. Email: {\tt\footnotesize lingqiao.liu@adelaide.edu.au}
}}

\begin{document}

\maketitle
\thispagestyle{empty}
\pagestyle{empty}

\begin{abstract}

Generalist robot manipulation policies (\GMP) have the potential to generalize across a wide range of tasks, devices, and environments.  
However, existing policies continue to struggle with out-of-distribution scenarios due to the inherent difficulty of collecting sufficient action data to cover extensively diverse domains.
\revision{While fine-tuning offers a practical way to quickly adapt a \GMP~to novel domains and tasks with limited samples, we observe that the performance of the resulting \GMP~differs significantly with respect to the design choices of fine-tuning strategies.}
\revision{In this work, we first conduct an in-depth empirical study to investigate the effect of key factors in \GMP~fine-tuning strategies, covering the action space, policy head, supervision signal and the choice of tunable parameters, where \textit{2,500 rollouts} are evaluated for \textit{a single configuration}. We systematically discuss and summarize our findings and identify the key design choices, which we believe give a practical guideline for \GMP~fine-tuning.}
\revision{We observe that in a low-data regime, with carefully chosen fine-tuning strategies, a \GMP~significantly outperforms the state-of-the-art imitation learning algorithms.} \revision{The results presented in this work establish a new baseline for future studies on fine-tuned \GMP, and provide a significant addition to the \GMP~toolbox for the community.}

\end{abstract}

\section{INTRODUCTION}

Recently, generalist robot manipulation policies (GMPs) have gained substantial attention. Pretrained on extensive datasets and leveraging cutting-edge architectures, it aims to perform diverse manipulation tasks through a single, unified policy, emphasizing broad generalization.

\revision{The predecessors of GMPs train a policy using in-house collected datasets for constrained scenarios~\cite{brohan2022rt, brohanRT2VisionLanguageActionModels}.} Inspired by the success of the foundation vision~\cite{radford2021learning} and language models~\cite{achiam2023gpt} on pretrained on large-scale datasets, the Open X-Embodiment dataset~\cite{collaborationOpenXEmbodimentRobotic2023} was introduced to significantly boost the development of GMPs, \revision{which collected the data distributed across institutions and embodiments. Key to the unified protocol is the shared action mode, i.e., the delta end-effector poses. Built upon the Open X-Embodiment dataset, training large-scale cross-embodiment policies is made possible. Octo~\cite{ghoshOctoOpenSourceGeneralist} and OpenVLA~\cite{karamchetiPrismaticVLMsInvestigating2024} demonstrated the capabilities of the generalist language-conditioned policies by either training from scratch with Transformers~\cite{vaswani2017attention} or fine-tuning an existing LLM with adaptable features.}

Despite these advancements, GMPs still struggle to generalize across diverse scenarios, especially the ones unseen from training data distributions~\cite{liEvaluatingRealWorldRobot2024}. 
Previous generalization tests were typically conducted in environments closely resembling their pretraining setups \cite{ghoshOctoOpenSourceGeneralist, kimOpenVLAOpenSourceVisionLanguageAction2024}, with only minor variations, such as object placements or background changes. The inherent complexity of embodied robotics, encompassing variations in visual appearances, embodiments, and tasks, remains a significant challenge.

\begin{figure}
  \centering
\centerline{\epsfig{figure=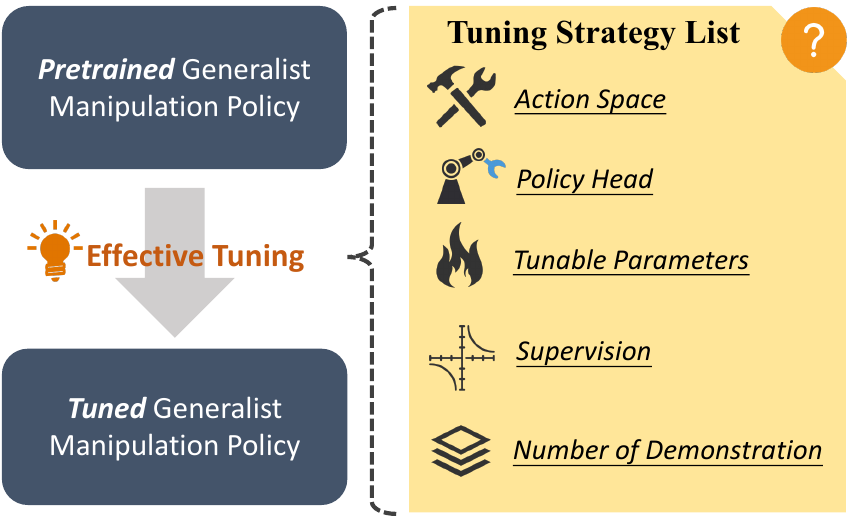,width=0.48\textwidth}}
\caption{For effective fine-tuning of generalist robot manipulation policies, a deeper understanding of several fundamental choices—action space, policy head, loss function, tunable parameters, and the number of demonstrations—is essential, as these seemingly basic elements exert a profound influence on tuned \GMP'~performance.}
\label{fig:Fig_video_SOD_motivation}
\vspace{-0.6cm}
\end{figure}

\begin{figure*}[hbtp!]

  \centering
\vspace{-0.5cm} \centerline{\epsfig{figure=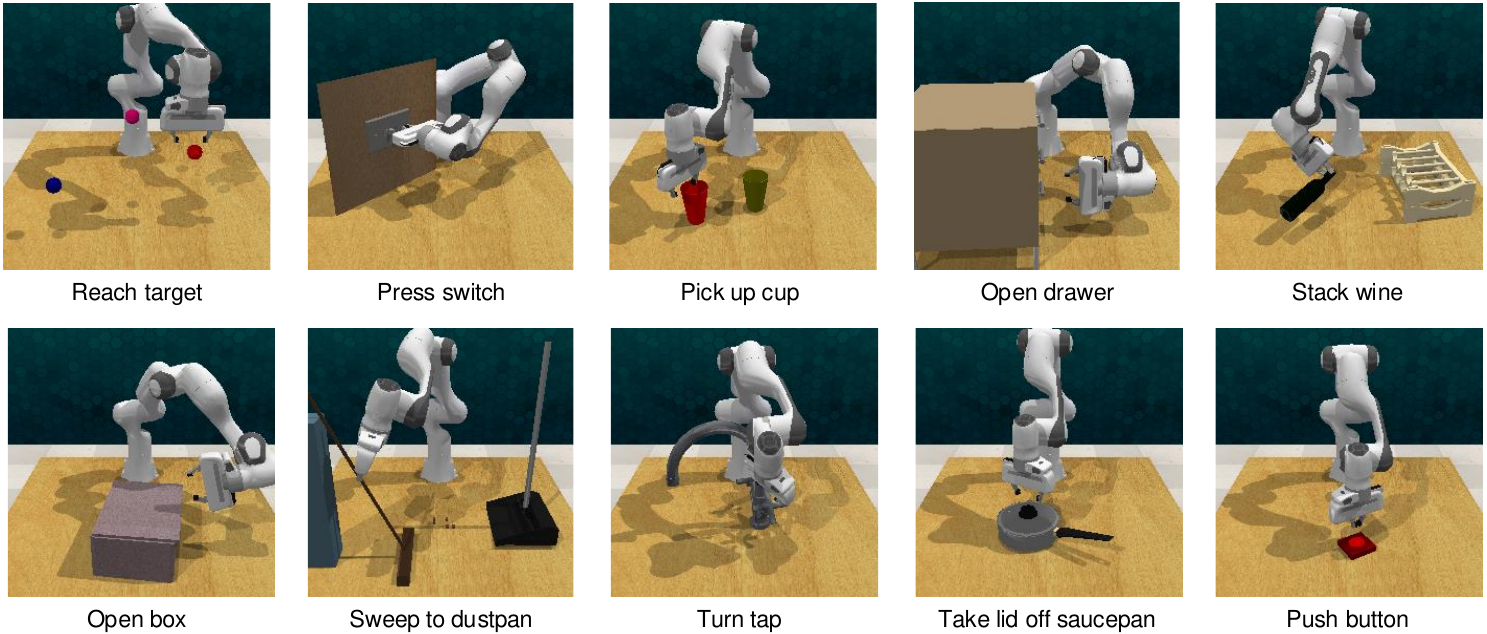,width=0.8\textwidth}}
 \caption{We systematically explore effective tuning strategies for generalist manipulation policies Octo~\cite{OctoOpenSourceGeneralist} on 10 tasks in the RLBench benchmark.}
\label{fig:rlbench}
	\vspace{-0.5cm}
\end{figure*}

\revision{While prior works have provided some initial investigations of the fine-tuning strategies~\cite{ghoshOctoOpenSourceGeneralist,karamchetiPrismaticVLMsInvestigating2024}, they still lack comprehensiveness on the following aspects:} 1) \textit{Underexplored Tuning Settings:} Key design choices, such as the choice of policy head~\cite{ghoshOctoOpenSourceGeneralist}, action space~\cite{mazzagliaRedundancyawareActionSpaces2024} etc., can significantly impact performance but were not systematically explored. Without a clear understanding of these fundamental settings, the development of more advanced fine-tuning methods may be hindered because we cannot fairly assess whether improvements arise from the proposed tuning method or from these underlying design choices. 2) \textit{Fixed Demonstration Quantities:} The number of demonstrations used during fine-tuning plays a pivotal role in the benefits of pretraining. Prior works fixed this quantity, failing to explore how varying it might influence performance. Without a clear understanding of the number of demonstrations at which \GMP~has its advantages, the direction for further improvements may become ambiguous. 3) \textit{Reliability and Reproducibility:}
Imitation learning in manipulation is highly unstable~\cite{mandlekarWhatMattersLearning2021}, with each training runs often producing inconsistent outcomes even with the same configuration. Previous studies~\cite{ghoshOctoOpenSourceGeneralist,karamchetiPrismaticVLMsInvestigating2024} relied on only a single training run and limited evaluation scale, raising concerns about the reliability of their insights.

\revision{In this work, instead of aiming to create a new GMPs framework, we focus on understanding the key design choices of GMPs through extensive empirical studies.}
Specifically, we systematically compare various fine-tuning techniques and design choices, including action space, policy head, tunable parameters, and \revision{supervision signals}.
\revision{We also investigate the demonstration quantities and how it impacts the performance of GMPs to understand the minimal demonstrations for GMPs to achieve good fine-tuning performances.}

In our experiments, \revision{to ensure statistical reproducibility, we adopt a unified evaluation setting using a widely adopted simulation platform, RLBench~\cite{jamesRLBenchRobotLearning2019}, and perform \textit{around 2,500 rollouts per each single configuration.}} Based on our findings, we show that a fine-tuned \GMP~significantly outperforms the state-of-the-art imitation learning algorithms, e.g., ACT~\cite{zhao2023learning} and Diffusion Policies~\cite{chiDiffusionPolicyVisuomotor2023} over all tasks in the low-data regime. We believe our work can provide some insights on tuning GMPs for future research.

\section{RELATED WORK}
\subsection{Generalist Robot Manipulation Policies}\label{subsec:gmp}
In recent years, there has been a growing focus on generalist robot manipulation policies. 
Early approaches such as RT-1~\cite{brohan2022rt} introduced a robotics Transformer to encode high-dimensional inputs and outputs and trained it on a broad dataset of various robotic tasks.
The subsequent work RT-2~\cite{brohanRT2VisionLanguageActionModels} utilized a larger vision-language backbone to create a more robust generalist policy. 
Then Open X-Embodiment~\cite{collaborationOpenXEmbodimentRobotic2023} compiled a large-scale real-world dataset by aggregating existing manipulation data, resulting in RT-1-X and RT-2-X, which retained the same architectures as RT-1 and RT-2 but achieve better performance. 
Based on the X-Embodiment dataset, Octo~\cite{OctoOpenSourceGeneralist} proposed a transformer-based policy with a modular attention structure, enabling flexible adaptation to new domains. OpenVLA~\cite{kimOpenVLAOpenSourceVisionLanguageAction2024} further advanced the field by extending training to the Open X-Embodiment dataset using LLama2~\cite{touvron2023llama}, resulting in a more powerful GMP. However, due to the model’s size of 7 billion parameters, even with the parameter-efficient tuning method LoRA~\cite{hu2021lora}, OpenVLA still requires substantial training resources.

Despite recent progress, the generalization ability of \GMP~remains unsatisfactory. 
These limitations are further revealed by a recently released benchmark SIMPLER~\cite{liEvaluatingRealWorldRobot2024}, which reconstructs simulated test environments from real-world scenes and adjusts them so that the simulation results closely align with real-world outcomes. 
This alignment allows researchers to conveniently evaluate the performance of GMPs in conditions that mirror real-world scenarios. For instance, SIMPLER reported that changing the texture of the robot arm results in a success rate drop of over 20\% for several GMPs. Additionally, viewpoint changes — a common out-of-distribution scenario~\cite{yuanLearningManipulateAnywhere2024} — can also be evaluated in SIMPLER\footnote{This phenomenon is observed by using the official codebase of SIMPLER}. We observed that GMPs often fail to complete tasks when the camera viewpoint changes, leading to erratic and uncontrolled arm movements.

\begin{table*}[htbp]
\centering
\caption{Benchmark of our model with state-of-the-art imitation learning methods, ACT and Diffusion Policy.}
\label{tab:benchmark_rlbench}
\renewcommand{\arraystretch}{0.70}
\begin{tabular}{l|cccccccccc|c}
\toprule
 \multirow{2}{*}{\textbf{Method}}  & \textbf{Reach} & \textbf{Press} & \textbf{Pick Up} & \textbf{Open} & \textbf{Stack} & \textbf{Open} & \textbf{Sweep} &\textbf{Turn} & \textbf{Push}& \textbf{Take} &  \multirow{2}{*}{\textbf{Avg.}} \\ 
 & \textbf{Target} & \textbf{Switch} & \textbf{Cup} & \textbf{Drawer} &\textbf{Wine} & \textbf{Box}& \textbf{Dust} & \textbf{Tap} & \textbf{Button} & \textbf{Lid} &\\ 
\midrule
\multicolumn{12}{c}{10 Demonstrations} \\
\midrule
Diffusion Policy~\cite{chiDiffusionPolicyVisuomotor2023}      & 0.0 & 0.0 & 30.0 & 0.0 & 1.8 & 0.0 & 0.0 & 0.0 & 36.8 & 100.0 & 16.9 \\
ACT~\cite{zhao2023learning}                    & 0.0 & 28.0 & 0.0 & 0.0 & 20.0 & 8.0 & 0.0 & 44.0 & 16.0 & 0.0 & 11.6 \\
\textbf{Ours }        & 17.8 & 30.0 & 7.8 & 41.0 & 6.0 & 58.8 & 30.0 & 7.0 & 10.0 & 12.0 & \textbf{22.0} \\
\midrule
\multicolumn{12}{c}{20 Demonstrations} \\
\midrule
Diffusion Policy~\cite{chiDiffusionPolicyVisuomotor2023}      & 0.0 & 0.0 & 0.0 & 0.0 & 0.0 & 0.0 & 0.0 & 0.0 & 100.0 & 39.4 & 13.9 \\
ACT~\cite{zhao2023learning}                    & 4.0 & 56.0 & 1.4 & 0.0 & 26.8 & 18.8 & 25.8 & 10.8 & 12.8 & 20.8 & 17.7 \\
\textbf{Ours}         & 16.0 & 34.4 & 16.0 & 45.8 & 14.4 & 70.0 & 48.0 & 19.4 & 22.8 & 28.0 & \textbf{31.5} \\
\midrule
\multicolumn{12}{c}{50 Demonstrations} \\
\midrule
Diffusion Policy~\cite{chiDiffusionPolicyVisuomotor2023}      & 0.0 & 36.8 & 70.4 & 0.0 & 85.2 & 44.4 & 3.6 & 11.2 & 7.6 & 100.0 & 36.0 \\
ACT~\cite{zhao2023learning}                    & 0.0 & 34.8 & 37.6 & 40.0 & 33.2 & 32.0 & 14.8 & 69.2 & 49.2 & 68.0 & 37.9 \\
\textbf{Ours}         & 48.4 & 50.0 & 24.0 & 58.4 & 42.4 & 94.4 & 48.0 & 37.6 & 58.8 & 64.8 & \textbf{52.7} \\
\midrule
\multicolumn{12}{c}{100 Demonstrations} \\
\midrule
Diffusion Policy~\cite{chiDiffusionPolicyVisuomotor2023}      & 56.0 & 18.4 & 13.6 & 5.2 & 6.0 & 49.2 & 6.0 & 5.6 & 32.4 & 32.8 & 22.5 \\
ACT~\cite{zhao2023learning}                    & 45.3 & 18.8 & 62.8 & 81.2 & 68.0 & 88.0 & 32.0 & 97.2 & 78.8 & 73.2 & 64.5 \\

\textbf{Ours}         & 60.0 & 46.0 & 44.0 & 84.0 & 60.8 & 94.8 & 69.2 & 28.0 & 83.2 & 75.2 & \textbf{64.6} \\
\bottomrule
\end{tabular}
\vspace{-10pt}
\end{table*}

\subsection{Transfer Learning in Manipulation Tasks}
In recent years, transfer learning has become increasingly popular in robotics research~\cite{jaquierTransferLearningRobotics2023} as it alleviates the cost of collecting new data for each manipulation task. Previous works focus broadly on the simulation-to-real transfer learning~\cite{arndt2020meta}, task transfer learning~\cite{james2018task,bonardi2020learning} and cross-embodiment transfer learning~\cite{yang2024pushing,chen2024mirage}. Orthogonal to prior works, we do not focus on a specific type of domain shift, whether it occurs in the environment, embodiment, or tasks. our approach targets transferring a \GMP~to a new domain, \wb{in more general context.}
In fact, the original Octo paper has reported basic fine-tuning experiments, comparing its performance with~\cite{majumdar2023we} and a transformer-based model trained from scratch across four tasks. OpenVLA provided a more comprehensive exploration, testing across seven tasks and comparing the results with the Diffusion Policy. While these works explored important aspects, we raised concerns in the introduction, and in this paper, we conduct a more systematic investigation.

\section{PRELIMINARIES}

\hsy{
We choose Octo-Small \cite{ghoshOctoOpenSourceGeneralist} as our model for fine-tuning \GMP~because it represents a transformer-based structure, which is a (almost) standard in \GMP. It achieves a balance of efficiency and success rate, with a size that provides enough parameters for \GMP~without being too large for extensive evaluation. Octo-Small uses ViT-S~\cite{dosovitskiy2020image} as its backbone and employs multiple convolution layers as its vision encoder, noting decreased performance with larger encoders like ResNet~\cite{he2016deep} on large-scale datasets. Although Octo includes a T5 language encoder~\cite{raffel2020exploring}, we have frozen it for this fine-tuning study. Octo introduces a novel diffusion-based policy head inspired by Diffusion Policy; we also explore a linear head, which empirically shows better performance \wb{during fine-tuning}.

The input to Octo includes tokens from the current RGB observation via a third-view camera, a fixed language token, and a specific action token. The action token is randomly initialized to aggregate features. These tokens are processed using attention layers with the causal mask. In the final layer, the action token is fed to the policy head to predict the final action. Originally, Octo uses a diffusion head with a \delee~action space, where the action dimension includes delta x, y, z, yaw, roll, pitch, and a gripper open/close action.
}

\begin{table*}[htbp]
\centering
\caption{Choice of Action Space: Delta Joint Position, Absolute Joint Position, Delta End-Effector, Absolute End-Effector. Success rates are expressed in percentages. The best average success rates (\textbf{Avg.}) are highlighted in bold.}
\renewcommand{\arraystretch}{0.8}
\label{tab:action_space}
\begin{tabular}{l|cccccccccc|c}
\toprule
 \multirow{2}{*}{\textbf{Action Space}}  & \textbf{Reach} & \textbf{Press} & \textbf{Pick Up} & \textbf{Open} & \textbf{Stack} & \textbf{Open} & \textbf{Sweep} &\textbf{Turn} & \textbf{Push}& \textbf{Take} &  \multirow{2}{*}{\textbf{Avg.}} \\ & \textbf{Target} & \textbf{Switch} & \textbf{Cup} & \textbf{Drawer} &\textbf{Wine} & \textbf{Box}& \textbf{Dust} & \textbf{Tap} & \textbf{Button} & \textbf{Lid} &\\ 
\midrule
\multicolumn{12}{c}{20 Demonstrations} \\
\midrule

\text{Absolute End-Effector}    &      0.0 &     0.0 &     0.0 &     0.0 &     0.0 &     0.0 &     0.0 &     0.0 &    0.0 &     0.0 &  0.0 \\ 
     Delta End-Effector &     0.0 &     0.0 &     3.6 &     0.0 &     0.0 &     0.0 &     0.0 &     0.0 &    12.8 &     41.6 &  5.8 \\
Absolute Joint Position &     4.4 &     0.4 &     5.6 &    14.4 &     1.6 &    20.4 &    27.2 &     4.4 &    13.6 &      7.2 &  9.9 \\
   Delta Joint Position &     6.8 &    14.4 &     8.8 &    34.8 &    13.2 &    68.0 &    42.4 &    10.4 &     7.2 &     11.6 & \textbf{21.6} \\
\midrule
\multicolumn{12}{c}{100 Demonstrations} \\
\midrule

\text{Absolute End-Effector}    &      0.0 &     0.0 &     0.0 &     0.0 &     0.0 &     0.0 &     0.0 &     0.0 &    0.0 &     0.0 &  0.0\\
     Delta End-Effector &    24.4 &     3.6 &    18.4 &     0.0 &     1.6 &    32.4 &    37.6 &    15.6 &    71.2 &     81.6 & 28.7 \\
Absolute Joint Position &    49.6 &    14.4 &    27.2 &    35.2 &    33.2 &    69.2 &    30.8 &     8.8 &    71.2 &     37.6 & 37.7 \\
   Delta Joint Position &    60.0 &    46.0 &    44.0 &    84.0 &    60.8 &    95.2 &    68.8 &    28.0 &    83.2 &     75.2 & \textbf{64.5} \\
\bottomrule
\end{tabular}
\vspace{-10pt}
\end{table*}

\section{Empirical Study Designs}\label{Emprical Study}

\subsection{What Factors to Be Investigated}
\hsy{We focus on how various design factors influence model performance. Selecting which factors to explore is challenging due to the many inherently coupled design elements. Our criteria are: 1) the \wb{choice of} factor cannot be easily determined by community consensus or existing research, but 2) \wb{has a strong potential to influence the performance.} We start with Octo’s default fine-tuning approach as a baseline, conducting small-scale experiments with various design choices. Promising designs are added to a pool list. From these, we derive effective tuning strategies and evaluate them through large-scale experiments. Hyperparameters like learning rate and batch size are beyond this scope; we perform a grid search to ensure training convergence.}

After a careful review of the design choices across various literature, as an initial study, we focus on the following aspects. 1) \textit{The action space representation.} Prior research have diverged action spaces when training GMPs. For example, it has been shown that absolute actions are important for training imitation learning policies with action chunking~\cite{chiDiffusionPolicyVisuomotor2023,zhao2023learning} with joint positions as the actions. However, for GMPs, Octo used delta end-effector poses as the actions~\cite{ghoshOctoOpenSourceGeneralist}. It remains unclear how to adapt a pretrained GMP to tasks that require flexible full-body control. \wb{2) \textit{Choice of policy head.}~Diffusion-based manipulation methods have achieved significant success in robot manipulation recently~\cite{chiDiffusionPolicyVisuomotor2023}. However, methods with diffusion designs have not been evaluated in fine-tuning scenarios. Considering that the training characteristics of the denoising process~\cite{ho2020denoising} differ from feedforward networks such as linear layers and multi-layer perceptron (MLP), a question arises: Do \GMP~with diffusion head still perform well in fine-tuning settings?} 3) \textit{Choice of tunable parameters.}~Normally when fine-tuning a pretrained model, people prefer either freezing the backbone or perform full-tuning~\cite{radford2021learning}. Nevertheless, full-tuning might cause overfitting to new data, and head-only fine-tuning might underfit the new data distribution. It remains unclear for a large pretrained GMP, what the best parameter tuning strategies should be. 4) \textit{Choice of supervision signals.}~
\wb{In light of the success of auxiliary loss in transfer learning~\cite{zhuang2020comprehensive} and recent attempts in manipulation tasks~\cite{ma2024actra}, we aim to explore auxiliary supervision for fine-tuning \GMP. In the fine-tuning setting, although the number of demonstrations is limited, each sample contains abundant action information. Specifically, an identical action can be represented in different ways, providing multiple labels as supervision signals. This naturally leads us to consider whether alternative action representations could be utilized as auxiliary supervision to assist tuning \GMP.}
Lastly, we aim to understand if the pretraining with large-scale data indeed allows GMPs to be more sample efficient during fine-tuning, than the existing state-of-the-art sample efficient imitation learning algorithms.

\subsection{Performance Measurement}
We adopt the success rate as the metric for our evaluation. A common phenomenon in manipulation imitation learning method is the instability of training where checkpoints can vary substantially in performance during training, even when performance appears to converge~\cite{mandlekarWhatMattersLearning2021}, and we find this issue is significant in our experiments. We observed that models trained over multiple runs frequently produce highly inconsistent outcomes. For instance, our model's average success rate is 31.2\% with 20 demonstrations. When considering the average success rate across all tasks using the lowest success rate, the value drops to 18.8\%, while the highest success rate reaches 49.2\%. This results in a gap of 30.4\% points between the best and worst performances. 

To address this issue and ensure statistical significance~\cite{andrychowiczWhatMattersPolicy2020} in our findings, we conduct 25 test trials across 10 tasks and tune the model 10 times, resulting in a total of 2,500 rollouts for each configuration. This experimental scale is significantly larger than that of OpenVLA (129 rollouts) and Octo (80 rollouts), leading to more robust statistical results. We average the outcomes across multiple tasks for better comparisons. \wb{Additionally, we fix the training and test sets for all experiments, including those involving our model, ACT~\cite{zhaoLearningFineGrainedBimanual2023a} and Diffusion Policy~\cite{chiDiffusionPolicyVisuomotor2023}, to ensure fair and consistent comparisons.}

\subsection{Implementation Details}
\hsy{Simulated environments are more suitable than real-world experiments for large-scale evaluation. We use RLBench, a widely adopted manipulation benchmark, which offers a diverse range of tasks and provides convenience for comparing policies. Although SIMPLER~\cite{liEvaluatingRealWorldRobot2024}, as we mentioned in~\ref{subsec:gmp}, is specifically designed to evaluate \GMP, it does not support our study mainly due to the lack of of training dataset.

We select 10 tasks with varying difficulty and action modes, as depicted in Fig.~\ref{fig:rlbench}. Demonstrations are automatically generated by the official codebase. To eliminate randomness, we collect 100 training set and 25 test set as offline data for each task in advance. For experiments with different demonstration quantities, training data are all sampled in a fixed order from the 100 samples. To enable large-scale experiments, we developed a codebase capable of parallelized training and testing, implementing several optimizations to enhance efficiency and scalability.

For observation, although multiple view camera can provide stereo vision, and lead to better performance generally~\cite{goyalRVTRoboticView2023a}\cite{goyalRVT2LearningPrecise2024}, we only use the front third-view camera as the input observation, as it has been reported~\cite{ghoshOctoOpenSourceGeneralist} to be more suitable for Octo compared to using multiple observations. This may be due to the lack of multi-observation data during pretraining. The number of fine-tuning iterations is determined through hyperparameter search and set to 3,000, which is sufficient to ensure convergence.

}

\begin{table*}[htbp]
\centering
\caption{Choice of policy head: Diffusion Head, Linear Head, MLP Head. Success rates are expressed in percentages. The best average success rates (\textbf{Avg.}) are highlighted in bold.}
\label{tab:action_head}
\renewcommand{\arraystretch}{0.8}
\begin{tabular}{l|cccccccccc|c}
\toprule
 \multirow{2}{*}{\textbf{Policy Head}}  & \textbf{Reach} & \textbf{Press} & \textbf{Pick Up} & \textbf{Open} & \textbf{Stack} & \textbf{Open} & \textbf{Sweep} &\textbf{Turn} & \textbf{Push}& \textbf{Take} &  \multirow{2}{*}{\textbf{Avg.}} \\ 
 & \textbf{Target} & \textbf{Switch} & \textbf{Cup} & \textbf{Drawer} &\textbf{Wine} & \textbf{Box}& \textbf{Dust} & \textbf{Tap} & \textbf{Button} & \textbf{Lid} &\\ 
\midrule
\multicolumn{12}{c}{20 Demonstrations} \\
\midrule
Diffusion Head &         14.0 &         16.0 &        18.0 &        24.0 &        0.0 &     40.0 &       10.0 &      6.0 &        16.0 &     18.0 & 16.2 \\
MLP           &         18.0 &         28.0 &        11.8 &        32.4 &        9.4 &     71.8 &       36.4 &     26.0 &        21.8 &     31.4 & 28.7 \\
Linear Head   &         10.4 &         36.4 &         5.8 &        43.8 &       16.0 &     76.4 &       47.8 &     22.4 &        14.8 &     39.8 & \textbf{31.4} \\
\midrule
\multicolumn{12}{c}{100 Demonstrations} \\
\midrule
Diffusion Head &         56.0 &         18.8 &        13.8 &         5.0 &        6.0 &     49.4 &        6.0 &      5.8 &        32.4 &     32.8 & 22.6 \\
MLP &55.6 &
46.8 &
38.4 &
66.8 &
48.8 &
90.8 &
61.4 &
29.6 &
78.8 &
75.4 &
59.2\\
Linear Head   &         60.0 &         46.0 &        44.0 &        84.0 &       61.0 &     95.0 &       69.0 &     28.0 &        83.0 &     75.0 & \textbf{64.5} \\
\bottomrule
\end{tabular}
\end{table*}

\begin{table*}[htbp]
\centering
\caption{Choice of Tunable Parameters: Full-tuning, Frozen Backbone, Head Tuning. Success rates are expressed in percentages. The best average success rates (\textbf{Avg.}) are highlighted in bold.}
\label{tab:tunable_parameters}
\renewcommand{\arraystretch}{0.8}
\begin{tabular}{l|cccccccccc|c}
\toprule
 \multirow{2}{*}{\textbf{Tunable Parameters}}  & \textbf{Reach} & \textbf{Press} & \textbf{Pick Up} & \textbf{Open} & \textbf{Stack} & \textbf{Open} & \textbf{Sweep} &\textbf{Turn} & \textbf{Push}& \textbf{Take} &  \multirow{2}{*}{\textbf{Avg.}} \\ 
 & \textbf{Target} & \textbf{Switch} & \textbf{Cup} & \textbf{Drawer} &\textbf{Wine} & \textbf{Box}& \textbf{Dust} & \textbf{Tap} & \textbf{Button} & \textbf{Lid} &\\ 
\midrule
\multicolumn{12}{c}{20 Demonstrations} \\
\midrule
Head Only      &         13.2 &         16.8 &         6.8 &         3.4 &        0.0 &     15.8 &        1.4 &     17.8 &        28.0 &      6.4 & 11.0 \\
Frozen Backbone &         24.8 &          9.4 &         5.4 &        20.0 &        2.8 &     41.8 &       12.8 &     28.0 &        14.4 &      7.4 & 16.7 \\
Full-tuning     &         10.4 &         36.4 &         5.8 &        43.8 &       16.0 &     76.4 &       47.8 &     22.4 &        14.8 &     39.8 & \textbf{31.4} \\
\midrule
\multicolumn{12}{c}{100 Demonstrations} \\
\midrule
Head Only      &          8.8 &         31.4 &         3.4 &        15.2 &        5.8 &     21.8 &        3.4 &      5.8 &        67.4 &      1.0 & 16.4 \\
Frozen Backbone &         53.4 &         39.8 &        20.0 &        38.8 &       34.4 &     42.8 &       73.4 &     18.4 &        63.4 &     38.4 & 42.3 \\
Full-tuning     &         60.0 &         46.0 &        44.0 &        84.0 &       61.0 &     95.0 &       69.0 &     28.0 &        83.0 &     75.0 & \textbf{64.5} \\
\bottomrule
\end{tabular}
\vspace{-10pt}
\end{table*}

\section{Summary of Findings}
Based on our empirical study, the findings highlight that the optimal configuration for action space, policy head, and tunable parameters is \textit{delta joint position, linear head, and full-parameter tuning}. Auxiliary supervision proves beneficial when data is limited, but becomes unnecessary with sufficient data. Ultimately, to address more general scenarios, we adopt the configuration—\textit{delta joint position, linear head, full-parameter tuning, and no auxiliary supervision}—which yields a model that outperforms both ACT and Diffusion Policy, as demonstrated in Tab~\ref{tab:benchmark_rlbench}. Below is a detailed analysis of these designs.
\subsection{Choice of Action Space: Delta Joint Position, Absolute Joint Position, Delta End-Effector, Absolute End-Effector}
\label{subsec:action_space}

\wb{To investigate the influence of action space, we compare the performance with \deljoint, \absjoint, \delee, \absee. The results are shown in Tab.~\ref{tab:action_space}, which demonstrate that the choice of action space significantly impacts performance. It shows that \deljoint~shows the best performance along 20 and 100 demonstrations.  

We attribute the good performance of \deljoint~to two folds: first, compared to \absjoint,  \deljoint~evidently has a smaller gap with pretraining, as both use the delta action space.
second, joint action space enable highly flexible control skill with the whole body control, without which \delee~often violates inverse kinematic constraints~\cite{mazzagliaRedundancyawareActionSpaces2024,maHierarchicalDiffusionPolicy2024}, leading to failures at most of time. As for \absjoint, it fails at all time when facing action space domain gap and inverse kinematic constraints simultaneously.\\\\
\textbf{Remark}~\deljoint~action space yielded the best performance. This findings contrasts with common practices in prior research~\cite{ghoshOctoOpenSourceGeneralist}, where the \delee~action space is adopted.}

\subsection{Choice of Policy Head: Diffusion Head, Linear Head, MLP Head}
\begin{table*}[htbp]
\renewcommand{\arraystretch}{0.8}
\centering
\caption{Choice of Supervision: w/ auxiliary supervision, w/o auxiliary supervision. Success rates are presented as percentages, with the best average success rates (Avg.) and standard error (Stderr) highlighted in bold.}
\label{tab:supervision}
\renewcommand{\arraystretch}{0.95}
\begin{tabular}{l|cccccccccc|c|c}
\toprule
 \multirow{2}{*}{\textbf{Supervision}}  & \textbf{Reach} & \textbf{Press} & \textbf{Pick Up} & \textbf{Open} & \textbf{Stack} & \textbf{Open} & \textbf{Sweep} &\textbf{Turn} & \textbf{Push}& \textbf{Take} &  \multirow{2}{*}{\textbf{Avg.}}& \multirow{2}{*}{\textbf{Avg. of Stderr}} \\ 
 & \textbf{Target} & \textbf{Switch} & \textbf{Cup} & \textbf{Drawer} &\textbf{Wine} & \textbf{Box}& \textbf{Dust} & \textbf{Tap} & \textbf{Button} & \textbf{Lid} &\\ 
\midrule
\multicolumn{13}{c}{20 Demonstrations} \\
\midrule
With Aux. Sup.      &         16.0 &         34.4 &        16.0 &        45.8 &       14.4 &     69.8 &       48.0 &     19.4 &        23.2 &     28.0 & \textbf{31.5} & \textbf{7.6} \\
Without Aux. Sup.   &         10.4 &         36.4 &         5.8 &        43.8 &       16.0 &     76.4 &       47.8 &     22.4 &        14.8 &     39.8 & 31.4 & 8.9 \\
\midrule
\multicolumn{13}{c}{100 Demonstrations} \\
\midrule
With Aux. Sup.      &         51.8 &         50.4 &        50.4 &        65.4 &       69.2 &     91.4 &       35.8 &     28.4 &        78.8 &     71.8 & 59.3 & \textbf{7.7} \\
Without Aux. Sup.   &         60.0 &         46.0 &        44.0 &        84.0 &       61.0 &     95.0 &       69.0 &     28.0 &        83.0 &     75.0 & \textbf{64.5} & 8.3 \\
\bottomrule

\end{tabular}
\vspace{-10pt}
\end{table*}

As a pretrained model, Octo has demonstrated that the diffusion head is superior to a naive linear head for \GMP, which aligns with expectations. This is because diffusion-based architectures are able to model multi-modal action distributions. However, in the fine-tuning context, our results contradict this. As shown in~\ref{tab:action_head}, the linear head significantly outperformed the diffusion head, achieving nearly double the accuracy in 20 demonstrations setting. Even with 100 demonstrations, the diffusion head’s performance remained limited. To rule out the possibility that this discrepancy stems from the diffusion head's larger parameters, potentially causing overfitting during fine-tuning, we conducted a control experiment using a deeper and wider MLP with more parameters than the diffusion head. The results indicate that the MLP only marginally underperformed compared to the linear head, suggesting that the performance gap is primarily due to the inherent structure of the diffusion head rather than the number of parameters. \\\\
\textbf{Remark}~We found that a linear layer outperformed the diffusion head, which is surprising given that the diffusion head~\cite{ghoshOctoOpenSourceGeneralist,chiDiffusionPolicyVisuomotor2023} generally surpasses both the linear head and MLP in most cases. This suggests that fine-tuning may present particular challenges for the diffusion head.
\subsection{Choice of Tunable Parameters: Full-tuning, Frozen Backbone, Head Tuning }
\label{subsec:tunable_param}

We investigated fine-tuning strategies focusing on tunable parameters, results are shown in Tab.~\ref{tab:tunable_parameters}. Head Only indicate only tune the policy head, i.e. linear head for our experiments. Frozen Backbone indicates to tune the vision encoder and policy head. 

Our experiments show that full fine-tuning outperformed partial fine-tuning approaches, whether using a frozen backbone or fine-tuning only the head. Specifically, full fine-tuning consistently yielded superior results, both with 20 samples and 100 samples. On the other hand, head-only fine-tuning performed the worst, suggesting that \GMP~faces a significant domain gap in the RLBench setting, making it difficult to learn robust representations through head tuning alone. Interestingly, the frozen backbone approach showed considerable improvement over head-only fine-tuning, despite only adding a multi-layer convolutional visual encoder as the tunable component. This comparison highlights that the domain gap in the visual layer is a key factor influencing \GMP’s generalization capabilities. Note that we do not compare parameter-efficient fine-tuning techniques here as it aims to tune large model at low computation cost, which we find does not work for tuning a small model like Octo.\\\\
\textbf{Remark}~We find that naive full-tuning produced the best results. Although a similar finding is observed in large-scale models such as OpenVLA \cite{kimOpenVLAOpenSourceVisionLanguageAction2024}, this is the first time such results are observed in a smaller \GMP. Besides, the gap between Head Only and Frozen Backbone indicates that domain differences in visual perception significantly impact the generalization ability of GMP.

\subsection{Choice of Supervision: Without Auxiliary Supervision, With Auxiliary Supervision}\label{subsec:supervision}

\wb{

We designed an experiment where \deljoint~is used as the primary supervision, while end-effector-related information (i.e., \delee, \absee, and the \textit{current state of the end-effector}) is incorporated as auxiliary supervision. The loss function adheres to the default L1 loss implemented in Octo's codebase. The rationale behind this design is that with three auxiliary losses, we can utilize the ample labels available to their full extent, as the end-effector action space has a smaller domain gap with the pretraining data. The results are shown in~\ref{tab:supervision}.  We additionally report the average stand error of the success rate as an indicator of training stability, aiming to underscore its potential. As shown, With 20 demonstrations, auxiliary supervision reduces variance to a level comparable to that achieved with 100 demonstrations. Furthermore, auxiliary supervision exhibits slightly better performance than the setting without it. However, when more data is available (e.g., 100 demonstrations), joint supervision can actually decrease performance.\\\\
\textbf{Remark} Our findings suggest that auxiliary supervision helps reduce accuracy variance in low-data regimes for tuning \GMP, as shown in Tab.~\ref{tab:supervision}. However, as the amount of training data increases, joint supervision tends to hinder performance, possibly due to conflicting signals from multiple losses when the model has sufficient data to learn from the primary action space alone. This presents a potential direction to address how to better utilize data.}

\subsection{Analysis of Sample Efficiency}\label{subsec:demonstration}

\begin{figure}\label{fig:num_demo}
  \centering
 \includegraphics[width=\linewidth]{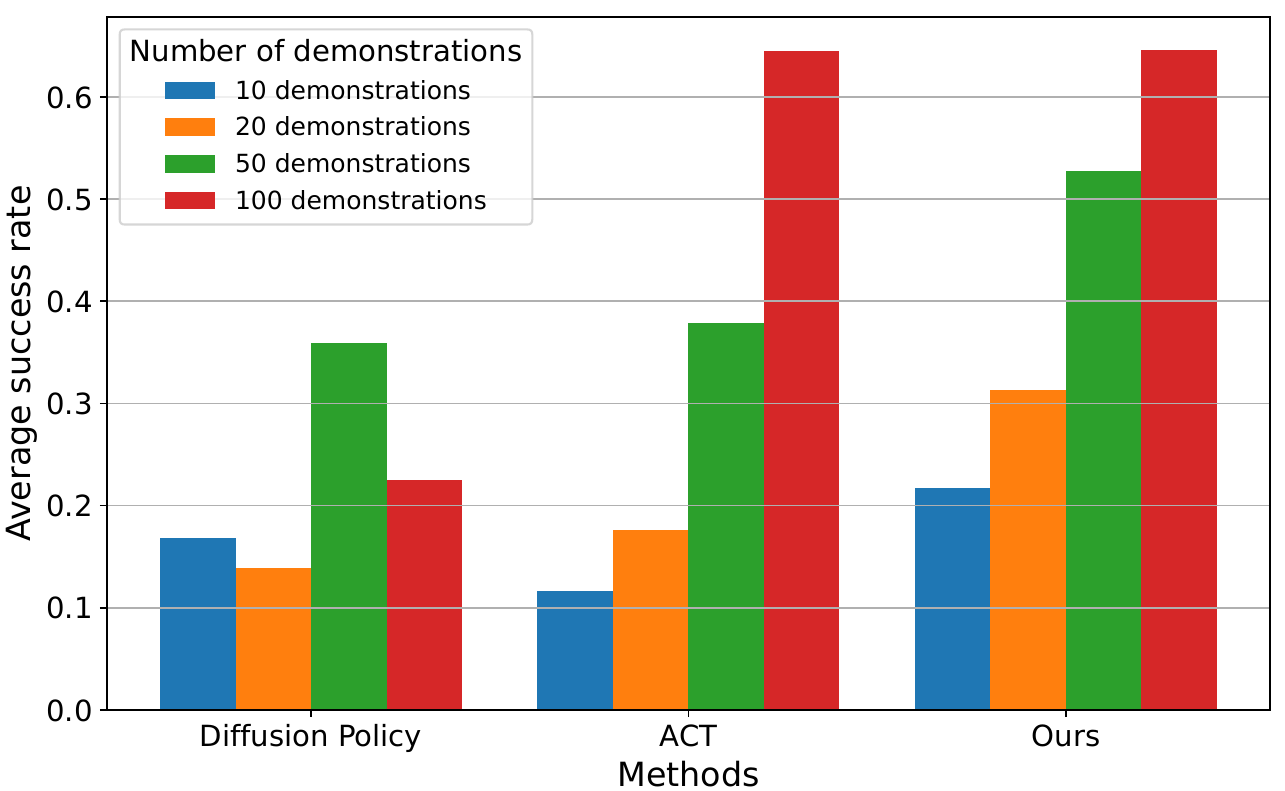}
 \vspace{-15pt}
 
\caption{Average success rate of Ours, ACT, Diffusion Policy across different demonstrations, details are in Tab.~\ref{tab:benchmark_rlbench}.}
\label{fig:benchmark_rlbench}
\vspace{-0.7cm}
\end{figure}

In this section, we vary the number of demonstrations used and observe the resulting accuracy trend, which is shown in Fig.~\ref{fig:benchmark_rlbench}. To better understand this trend, we also compare our results with ACT and Diffusion Policy. We found that with fewer demonstrations like 10, 20, and 50, the derived model consistently surpasses both as the number of demonstrations increases. But when the number of demonstrations reaches 100, the average accuracy of both methods becomes nearly identical. This phenomenon suggests that the pretrained parameters of \GMP~significantly aid model learning in low-data regimes, guiding it to converge towards a more generalizable parameter space. These findings indicate that the strength of fine-tuning \GMP~lies in few-shot learning scenarios, while in data-rich environments, its accuracy advantage may diminish.\\\\
\textbf{Remark}~\GMP~shows a clear advantage in low-data regime, outperforming ACT with fewer demonstrations. However, as the number of demonstrations increases, both methods perform similarly at 100 demonstrations. This highlights \GMP’s strength in low-data regimes, with diminishing advantages in data-rich settings.

\section{CONCLUSIONS}

In this work, we conducted a systematic empirical investigation into effective strategies for fine-tuning \GMP. We investigate several key design elements, including the action space, policy head, tunable parameters, and supervision signals. We systematically discuss and summarize our findings and identify the key design choices, which offers better understanding and a practical guideline for \GMP~fine-tuning. We also highlight the advantage of tuning \GMP~in low-data regimes, where it significantly outperforms state-of-the-art imitation learning algorithms. 
We believe these efforts will establish a new baseline for future studies and provide a valuable addition to the \GMP~toolbox for the community.

\newpage

\bibliographystyle{IEEEtran}
\bibliography{bibliography}

\end{document}